\documentclass[11pt,a4paper]{article}
\usepackage[hyperref]{eacl2021}
\usepackage{times}
\usepackage{latexsym}

\usepackage{microtype}

\usepackage{graphicx}
\usepackage{bbold}
\usepackage[symbol]{footmisc}

\aclfinalcopy 
\def\aclpaperid{17} 

\title{Graph Convolutional Network for Swahili News Classification}

\author{Alexandros Kastanos\footnotemark\\
  Independent \\
  \texttt{alecokastanos@gmail.com} \\\And
  Tyler Martin\footnotemark[1]\\
  Independent \\
  \texttt{tyler.a.martin12@gmail.com} \\}

\date{}

\begin{document}
\maketitle
\setcounter{footnote}{1}
\footnotetext{Equal contribution}
\begin{abstract}
This work empirically demonstrates the ability of Text Graph Convolutional Network (Text GCN) to outperform traditional natural language processing benchmarks for the task of semi-supervised Swahili news classification. In particular, we focus our experimentation on the sparsely-labelled semi-supervised context which is representative of the practical constraints facing low-resourced African languages. We follow up on this result by introducing a variant of the Text GCN model which utilises a bag of words embedding rather than a naive one-hot encoding to reduce the memory footprint of Text GCN whilst demonstrating similar predictive performance.
\end{abstract}

\section{Introduction}
\label{sec:introduction}
Text classification is a widespread natural language processing (NLP) task with applications including topic classification \cite{wang-manning-2012-baselines}, content moderation \cite{bodapati-etal-2019-neural}, and fake news detection \cite{wang-2017-liar}. News categorisation, another example application, is of particular relevance as it is used to automatically handle the incredible volume of news published daily. Effective classification assists in better article recommendation for readers, semantic topic categorisation, and removal of abusive or untrue content \cite{minaee-2020-deep}.

Traditional news classification techniques involve extracting features from the text followed by a classifier to generate predictions. More recently, large transformer-based pre-trained models have dominated text classification benchmarks in high-resource settings \cite{NEURIPS2019_XLNet,sun2020finetune}.

Graph Neural Networks (GNNs), a family of model architectures designed to operate directly on irregularly structured graphs, have seen a recent uptick in popularity in several applications. These include citation networks \cite{Fey_2018_CVPR, velickovic2018graph, KipfW17}, semantic role labelling \cite{marcheggiani-titov-2017-encoding}, machine translation \cite{beck-etal-2018-graph}, and named entity recognition \cite{cetoli-etal-2017-graph}. Although news texts superficially present a sequence of words, a document contains an implicit graph structure in the form of semantic and syntactic relationships \cite{mihalcea-tarau-2004-textrank}. A corpus can be arranged into a graph by making use of both inter-document and intra-document relationships. \citet{YaoM019} construct a single graph from the complete corpus while \citet{huang-etal-2019-text} construct a graph on a per-document basis.

\section{Motivation}
\label{sec:motivation}

Despite the importance of this application area and Swahili being the fourteenth most widely spoken language in the world \cite{eberhard}, there is a deficit of published work on text classification for Swahili documents. Some of the factors contributing to this under-representation include a shortage of freely-available annotated datasets, and limited literature comparing methods commonly applied to high-resource languages in a low-resource context \cite{abs-2003-11529, niyongabo-etal-2020-kinnews}. \citet{joshi-etal-2020-state} classify Swahili as a \textit{hopeful} language, indicating that although research endeavours and community-driven efforts to digitise and annotate data exist \cite{hcs-a-v2_en, davis_david_2020_4300294, SHIKALI2020105951}, there remains a sizeable gap between the NLP tools available for Swahili and high-resource languages.

Off the back of the effectiveness of transformer-based models in high-resource settings, efforts have been made to apply these architectures to low-resource languages. While multilingual models \cite{lample2019cross, conneau-etal-2020-unsupervised} have shown promising zero-shot cross-lingual results, the performance has been shown to vary greatly by target languages, often with a large drop for tasks in low-resource languages like  Swahili \cite{jiang-etal-2020-x, hu-2020-xtreme}. An alternative approach, with comparable results, involves transferring monolingual models to a target language \cite{tran2020english,artetxe-etal-2020-cross}. This method still relies on sequentially fine-tuning transformer models, which has financial and ethical implications \cite{strubell-etal-2019-energy}. In addition to these shortcomings, transformer-based methods typically impose a memory requirement which scales quadratically with the sequence length. Despite work to reduce this drawback \cite{zaheer2021big, beltagy2020longformer}, large transformer-based models remain computationally challenging in the context of African research and industry.

Consequently, we aim to combat these challenges with the following set of contributions:
\begin{itemize}
\item Provide a set of accessible traditional NLP benchmarks for Swahili news classification.
\item Empirically compare these benchmarks to a Text Graph Convolutional Network (Text GCN) solution which was initially developed for English. To our knowledge this is the first instance of a GNN being used for text classification on any African language dataset.
\item Focus experiments on the semi-supervised context where the number of labels are highly constrained and the compute is restricted.
\end{itemize}

\section{Graph Neural Networks}
\label{sec:gnn}
\begin{figure}[htbp]
    \centering
    \includegraphics[width=6cm]{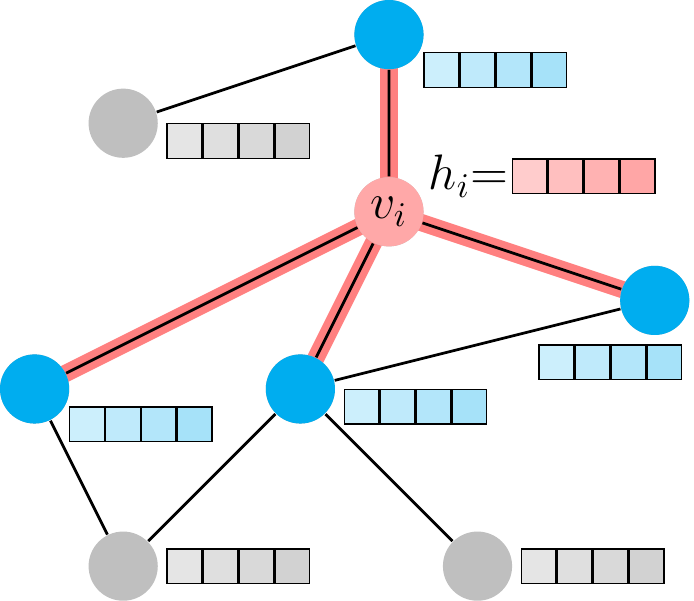}
    \caption{Information is pooled along the shaded edges in the graph and aggregated at node $v_i$ to update the hidden representation of $h_i$.}
    \label{fig:graph}
\end{figure}

Graph Neural Networks are a neural network model architecture which is able to generalise to non-euclidean data structures \cite{Gori_1555942, Scarselli_4700287,abs-1806-01261}. As with other modern deep learning architectures, a GNN is constructed by stacking a number of GNN layers sequentially. In fact, \citet{Wu_GNN_9046288} and \citet{zhou2018graph} present GNNs as a generalisation of convolutional neural networks. The underlying idea is that for a given graph $\mathcal{G}$, with nodes $\mathcal{V}$ and edges $\mathcal{E}$, one can learn a rich representation for each input node by aggregating information from its neighbourhood. Each node $v_i$ is initially represented by an embedding vector $x_i \in \mathbb{R}^F$ while the edge structure is stored in an adjacency matrix $A \in \mathcal{R}^{N \times N}$. $N$ is the number of nodes in $\mathcal{G}$. By stacking GNN layers, one can learn increasingly rich representations by incorporating features from neighbours of neighbours. Figure \ref{fig:graph} illustrates how the hidden representation of $h_i \in \mathcal{R}^{F'}$ is obtained by aggregating the 1-hop neighbourhood along the red shaded edges.

\subsection{Text Graph Convolutional Networks}
\label{sec:TextGCN}
Graph Convolutional Networks (GCNs), a particular type of GNN, aim to learn a localised and fast approximation of a spectral graph convolution \cite{KipfW17}. A GCN layer can be formalised mathematically as given in equation \ref{eq:gcn} where $H^{(\ell)} \in \mathbb{R}^{N \times F}$ is the node representation at layer $\ell$, $\Theta^{(\ell)} \in \mathbb{R}^{F \times F'}$ are trainable parameters, and $\phi(\cdot)$ is an element-wise activation function. The $\widetilde{D}^{-\frac{1}{2}} \widetilde{A} \widetilde{D}^{-\frac{1}{2}}$ term is the result of the \textit{renormalisation trick} wherein $\widetilde{A} = A + \mathbb{1}_N$ and $\widetilde{D}_{ii} = \sum_j \widetilde{A}_{ij}$.

\begin{equation} \label{eq:gcn}
    H^{(\ell + 1)} = \phi \left( \widetilde{D}^{-\frac{1}{2}} \widetilde{A} \widetilde{D}^{-\frac{1}{2}}  H^{(\ell)} \Theta^{(\ell)} \right)
\end{equation}

In semi-supervised classification, a GCN is trained using gradient descent where the loss is calculated using cross entropy over the subset of labelled nodes. Unlike many self-training approaches, which rely on sudo-labels generated by a supervised model trained on the labelled subset of the data \cite{jo-cinarel-2019-delta}, a GCN operates directly on all nodes and without the need to generate potentially noisy sudo-labels.

Text GCN applies this model to a corpus of documents by treating documents and words as nodes in a heterogeneous graph \cite{YaoM019}. Furthermore, Text GCN constructs a weighted adjacency matrix by representing document-word interactions using their TF-IDF value and word-word interactions using Positive Point-wise Mutual Information (PPMI) over a fixed length context window (see Appendix \ref{sec:textgcn_adj}).

\section{Experiments}
\label{sec:experiments}
All experiments presented below are conducted in a transductive setting and assume a fixed corpus size. Furthermore, results are presented in terms of the mean and standard deviation obtained by repeating all experiments 5 times with different seeds.
    
\subsection{Data}
\label{sec:data}
The Swahili News Classification dataset \cite{davis_david_2020_4300294} is used to compare the ability of each model to categorise news articles into one of six categories: \textit{kitaifa} (``national''), \textit{michezo} (``sports''), \textit{burudani} (``entertainment''), \textit{uchumi} (``economy''), \textit{kimataifa} (``international''), and \textit{afya} (``health''). In total, the data contains 23,266 labelled samples\footnote{We exclude a sample from the dataset which comprised of the text \texttt{[`.']}.} which we divide into train, validation, and test sets using a 8:1:1 split. Table \ref{tab:dataset} details the number of samples from each news category present in each subset. It is important to note the considerable class imbalance, which motivates the use of macro $F_1$ score as the primary metric in the results to follow.

\begin{table}
\centering
\begin{tabular}{crrr}
\hline \textbf{Class} & \textbf{Train} & \textbf{Validation}  & \textbf{Test} \\ \hline
kitaifa   & 8,193 & 1024 & 1025 \\
michezo   & 4,802 &  601 &  600 \\
burudani  & 1,783 &  223 &  223 \\
uchumi    & 1,622 &  202 &  203 \\
kimataifa & 1,525 &  191 &  190 \\
afya      &   687 &   86 &   86 \\
\hline
\textbf{Total} & 18,612 & 2,327 & 2,327 \\
\hline
\end{tabular}
\caption{\label{tab:dataset} The number of documents per class in the train, validation, and test subsets. }
\end{table}

The corpus undergoes a text processing stage which includes stemming using the SALAMA language manager \cite{Hurskainen2004SALAMA,Hurskainen1999SALAMASL} (see Appendix \ref{sec:data_processing}). Since code-switching is common in Swahili speakers \cite{ndubuisi-obi-etal-2019-wetin} and often degrades model performance \cite{piergallini-etal-2016-word}, we attempt to estimate the proportion of English tokens in the dataset. Without removing proper nouns or words that are shared between Swahili and English, an upper threshold estimate indicates that at most 2.41\% of all the words are code-switched. We deem this to be small enough that no special treatment is applied to account for code-switched tokens.

\subsection{Baselines}
\label{sec:baselines}
A set of traditional baseline models are compared to the graph-based solutions. In all cases the feature vectors, $X \in \mathcal{R}^{N \times 300}$, produced by these models are fed into a logistic regression classifier.

The first of these is the term frequency inverse document frequency (TF-IDF) model, which is a normalised representation of the relationship between word frequency and a particular document. The unnormalised version is the \textit{Counts} model, which simply uses the word count per document. The third baseline averages all word embeddings in the document to generate a document embedding. We use the pre-trained 300 dimension Swahili \textit{fastText} embeddings without bigrams \cite{bojanowski-etal-2017-enriching}. The final two baseline models are the PV-DBOW and PV-DM variants of the popular \texttt{doc2vec} model \cite{pmlr-v32-le14}. The former uses a distributed bag of words technique while the latter uses distributed memory. 

\subsection{Implementation}
\label{sec:implementation}
We implement two Text GCN variants. The first is the vanilla model where the input feature vectors are simply represented using one-hot vectors (as presented in \citet{YaoM019}), while the second variant, Text GCN-t2v (\texttt{text2vec}), makes use of \texttt{word2vec} and \texttt{doc2vec} embeddings to represent the word and document input features respectively. Both the \texttt{word2vec} and \texttt{doc2vec} representations were trained using the default parameter settings as per the original papers, with exception of a 20 epoch training limit. Both graph models make use of two GCN layers, each with a dropout rate of 0.5. The first layer applies a ReLU activation to the output from a 200-dimension hidden layer while the final layer applies the softmax operation over the output layer. The Adam optimiser \cite{KingmaB14} is used with a learning rate of 0.02 and the models are trained for a maximum of 100 epochs. Unless otherwise indicated, all PPMI weights in the graph are constructed using a window size of 30 and only 20\% of the training set nodes are labelled. The code to reproduce all experiments can be found online\footnote{\url{https://github.com/alecokas/swahili-text-gcn}}.

\subsection{Results and Discussion}
\label{sec:results}

Table \ref{tab:acc_results} presents the test set accuracy and macro $F_1$ score for all baseline and Text GCN models. Of the baseline models, we notice that the PV-DBOW and \textit{Counts} models perform the best in terms of $F_1$, while the averaged fastText vectors clearly perform worst. Both GCN variants outperform the traditional baselines on both test set metrics, with a mean $F_1$ score of 75.29\% and 75.67\% for Text GCN and Text GCN-t2v respectively. Although Text GCN-t2v does not significantly outperform Text GCN, it has a reduced memory footprint which makes it computationally more attractive. Reducing the input feature size reduces the training time and cloud costs by factors of 5 and 20 respectively (see Appendix \ref{sec:cloud}).
It is worth noting the discrepancy between the accuracy and $F_1$ metric, particularly for the TF-IDF and PV-DM baselines. In the remainder of our experiments, we focus on $F_1$ score as we are interested in models which are robust to class imbalance.

\begin{table}
\centering
\begin{tabular}{lcc}
\hline \textbf{Model} & \textbf{Accuracy (\%)} & \textbf{$\mathbf{F_1}$ (\%)} \\ \hline
TF-IDF                & 83.07 $\pm$ 0.00       &  68.72 $\pm$ 0.00             \\
Counts                & 83.32 $\pm$ 0.00       &  73.60 $\pm$ 0.00             \\
fastText              & 67.47 $\pm$ 0.00       &  32.41 $\pm$ 0.00              \\
PV-DBOW               & 81.64 $\pm$ 0.47       &  72.93 $\pm$ 0.75              \\
PV-DM                 & 77.01 $\pm$ 0.38       &  67.50 $\pm$ 0.64               \\
\hline
Text GCN              & 84.62 $\pm$ 0.10  & 75.29 $\pm$ 0.52  \\
Text GCN-t2v          & \textbf{85.40 $\pm$ 0.22}       & \textbf{75.67 $\pm$ 0.90}   \\
\hline
\end{tabular}
\caption{\label{tab:acc_results} Comparison of the mean and standard deviation test set accuracy and $F_1$ scores for all models.}
\end{table}

\begin{figure}[htbp]
    \centering
    \includegraphics[width=\columnwidth]{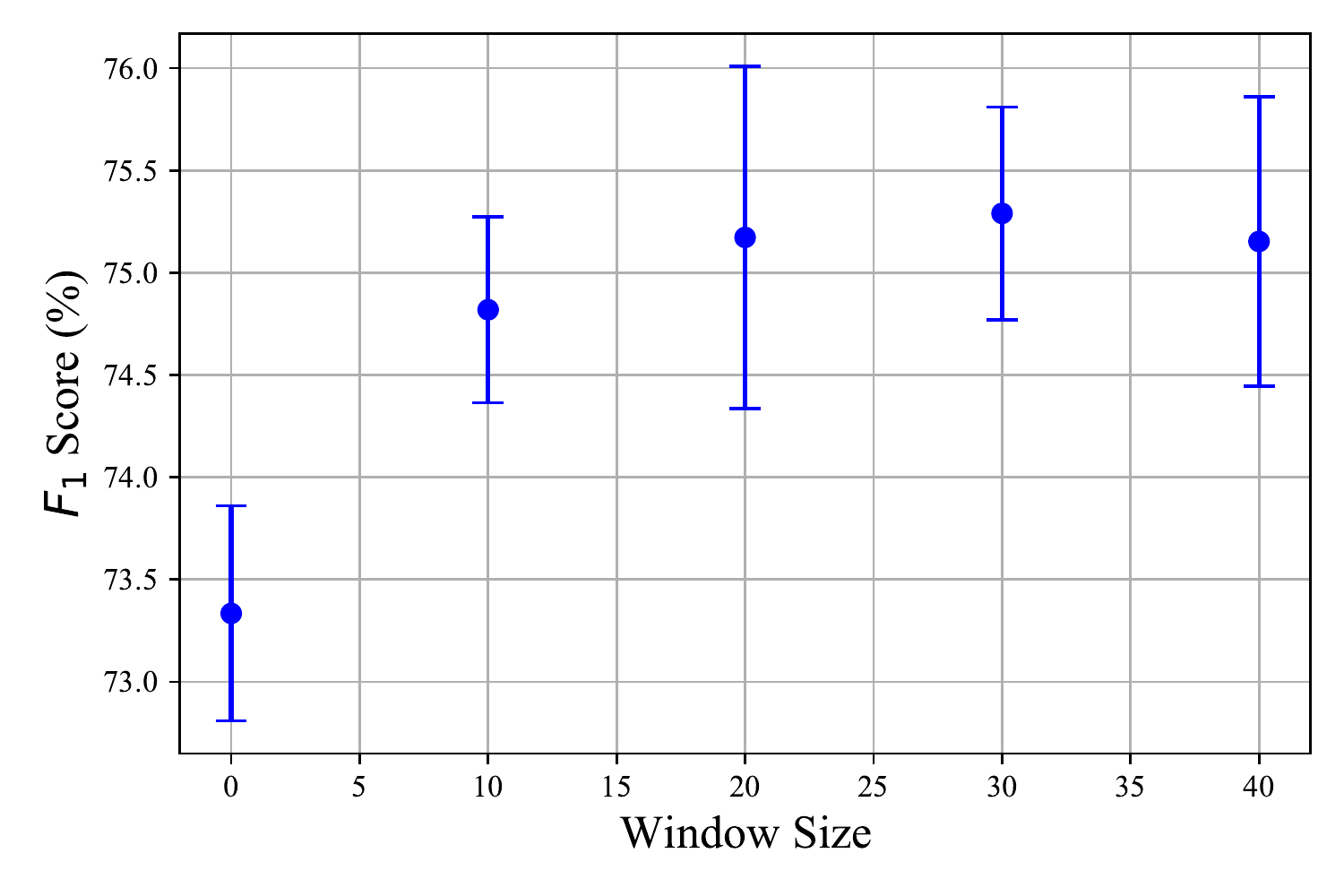}
    \caption{Text GCN performance for varied word co-occurrence window sizes.}
    \label{fig:windows}
\end{figure}

Figure \ref{fig:windows} examines the impact of the PPMI window size on the performance of Text GCN. These experiments demonstrate that for this Swahili news corpus, a window size of at least 20 is recommended to provide a context large enough to capture useful word co-occurrence statistics. Furthermore, there is a sharp drop in average $F_1$ performance to 73.33\% if PPMI is omitted entirely.

\begin{figure}[htbp]
    \centering
    \includegraphics[width=\columnwidth]{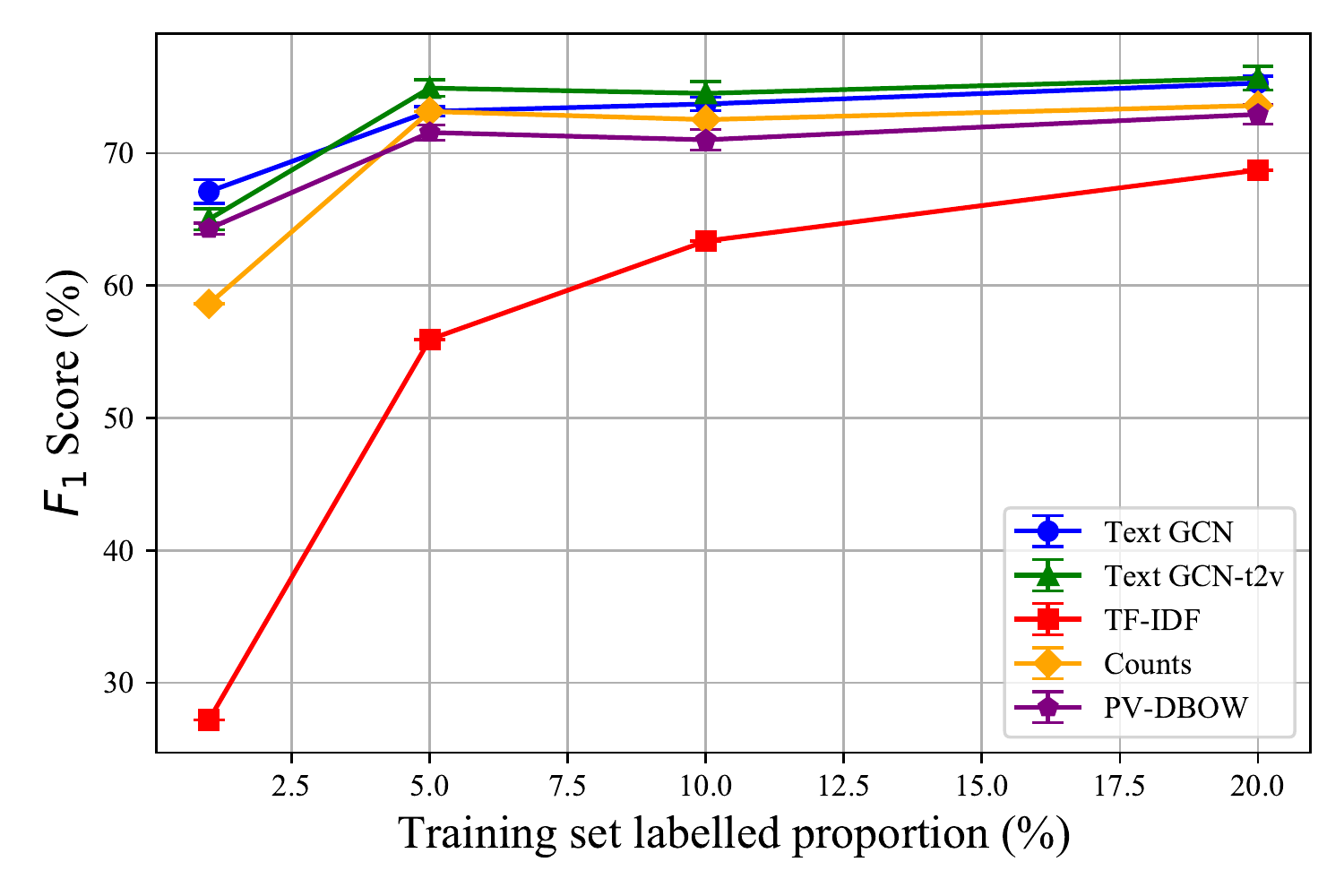}
    \caption{Test set macro $F_1$ scores using varying percentages of the training set labels.}
    \label{fig:proportions}
\end{figure}

As mentioned in section \ref{sec:motivation}, annotated data is particularly difficult to source for Swahili applications. Therefore, it is important to determine the effectiveness of the top performing models when the proportion of labels in the training set is drastically reduced. To this end, figure \ref{fig:proportions} compares the macro $F_1$ scores obtained by both Text GCN variants, TF-IDF, and the PV-DBOW models for training set proportions of 1\%, 5\%, 10\%, and 20\%. We find that both Text GCN models consistently outperform the traditional techniques, most noticeably as the proportion of labels is reduced. The TF-IDF and \textit{Counts} models in particular see a significant degradation in performance when reducing the label proportion from 5\% to 1\%, while PV-DBOW consistently lags behind both Text GCN models.

\section{Conclusion}
\label{sec:conclusion}
This paper empirically demonstrates the ability of Text GCN, a model originally developed for English, to outperform traditional models for the task of semi-supervised Swahili news classification. In doing so, we present an accessible set of baselines and demonstrate that when the number of training set labels is reduced, these methods fail to maintain their predictive ability. Finally, our Text GCN-t2v variant imposes a significantly reduced memory cost while continuing to match the predictive performance of the vanilla Text GCN. 

Our hope is that these results highlight the applicability of GNNs to semi-supervised Swahili applications more broadly than news categorisation. Future endeavours could extend this work to Swahili speech recognition where decoded outputs are often represented as a directed graph \cite{Ragni,Kastanos}.

\section*{Acknowledgments}
We thank Mario Ausseloos, Jacob Deasy, and Devin Taylor for their feedback on the draft manuscript. The authors would also like to thank the reviewers for their helpful comments and directions for future work.

\bibliography{anthology,eacl2021}
\bibliographystyle{acl_natbib}

\newpage
\appendix
\section{Text GCN Adjacency Matrix}
\label{sec:textgcn_adj}
A formal definition of the adjacency matrix, as applied in Text GCN, is provided in equation \ref{eq:adj} \cite{YaoM019}.
\begin{equation} \label{eq:adj}
    A_{i j}=\left\{
        \begin{array}{ll}
            \textnormal{PMI}(i, j) & i, j \textnormal{ are words; } \textnormal{PMI}(i, j)>0 \\
            \textnormal{TF-IDF}_{i j} & i \textnormal{ is a document, } j \textnormal{ is a word} \\
            1 & i=j \\
            0 & \textnormal{else}
        \end{array}\right.
\end{equation}

Through the constraint defined in the first condition, we implicitly interpret word co-occurrence as a Positive Point-wise Mutual Information (PPMI).
\begin{equation} \label{eq:PPMI}
    \textnormal{PPMI}(i, j) = \textnormal{max} \left( \textnormal{PMI}(i, j),\ 0 \right)
\end{equation}

Using $W(i)$ to indicate the number of sliding windows in which word $i$ occurs and $\#W$ as the total number of sliding windows, we can formulate PMI as follows:

\begin{equation} \label{eq:PMI}
    \textnormal{PMI}(i, j) = log \left( \frac{p(i,j)}{p(i)p(j)} \right)
\end{equation}

where the joint and marginal probabilities are given by equations \ref{eq:joint} and \ref{eq:marginal} respectively. 

\begin{equation} \label{eq:joint}
    p(i,j) = \frac{W(i,j)}{\textnormal{\#W}}
\end{equation}

\begin{equation} \label{eq:marginal}
    p(i) = \frac{W(i)}{\textnormal{\#W}}
\end{equation}

It is worth noting that while applying the \textit{renormalisation trick} mentioned in section \ref{sec:TextGCN}, we replace the original definition of $\widetilde{A}$ where $\widetilde{A} = A + \mathbb{1}_N$, with $\widetilde{A} = A$. This is because the identity diagonal is handled in equation \ref{eq:adj}.

\section{Data Processing}
\label{sec:data_processing}
This section describes the preprocessing pipeline applied to the Swahili News Dataset before training any of the baseline models or constructing the graph for the Text GCN models.

First, we drop one of the samples from the initial dataset as it simply contains the string `[.]'. A cleaning stage is applied to all remaining samples wherein all text is converted to lower case, Unicode characters are mapped to an ASCII equivalent, some Twitter meta information is removed, and superfluous whitespace characters are stripped.

Next, we tokenize the text into words using \texttt{word\_tokenize} from the NLTK library \cite{bird-2006-nltk}, and exclude stop words, single character words, and words containing non-alphabetical characters. All words which occur more than once are stemmed using the SALAMA language manager \cite{Hurskainen1999SALAMASL,Hurskainen2004SALAMA} and words longer than 30 characters are discarded. Finally, we use regular expressions to detect and merge onomatopoeic and laughter tokens to the special tokens \texttt{onomatopoeia} and \texttt{laughter} respectively.

\section{Cloud Compute Comparison}
\label{sec:cloud}
This Appendix provides further details on the training comparison between the vanilla Text GCN, which uses one-hot encoding to represent nodes, and the Text GCN-t2v variant, which uses \texttt{word2vec} and \texttt{doc2vec} embeddings to represent the word and document nodes respectively (hence \texttt{text2vec}).

Table \ref{tab:aws} compares two cloud machines from Amazon Web Services (AWS). Both machines are CPU only and are billed according to the total time the instance is running. The Text GCN model, with input features $X \in \mathcal{R}^{N \times N}$ requires over $32\ \textnormal{GB}$ of RAM during training, and therefore is trained on the \texttt{r5a.2xlarge} machine. On the other hand, the Text GCN-t2v has a radically reduced input feature space $X \in \mathcal{R}^{N \times 300}$, and therefore requires less than $16\ \textnormal{GB}$ of RAM. This allows us to train it on the significantly cheaper \texttt{r5a.large} machine.

\begin{table}[h!]
\centering
\begin{tabular}{ccc}
\hline 
\textbf{Machine Name} & \textbf{RAM (GB)} & \textbf{\$ / Hour} \\ \hline
r5a.large             & 16                & 0.133               \\
r5a.2xlarge           & 64                & 0.532               \\
\hline
\hline
\end{tabular}
\caption{\label{tab:aws} AWS machine and price comparison. Information as of February 2021. }
\end{table}

Table \ref{tab:time} presents the billable cloud time required for each model to construct their respective graphs and train to 100 epochs on their respective instances. We note that although the more sophisticated node representation results in Text GCN-t2v taking 15 minutes longer to construct the graph representation, the resulting model trains far quicker that the Text GCN equivalent. As a result, Text GCN-t2v only incurs costs for just over an hour while Text GCN remains running for over 5 hours.

\begin{table}
\centering
\begin{tabular}{lrr}
\hline 
                & \textbf{Text GCN} & \textbf{Text GCN-t2v} \\ \hline
Build graph     & 30               & 45  \\
Train model     & 273              & 16  \\
\hline
\textbf{Total}  & \textbf{303}     & \textbf{61}  \\
\hline \hline
\end{tabular}
\caption{\label{tab:time} Cloud time comparison between Text GCN and Text GCN-t2v. All times are reported in minutes.}
\end{table}

Combining the information from tables \ref{tab:aws} and \ref{tab:time}, the total cost to train Text GCN is $\$2.69$ while the cost to train Text GCN-t2v is estimated at $\$0.14$ \footnote{Pricing listed at \url{https://aws.amazon.com/ec2/pricing/on-demand/} as of February 2021.}. This comparative cost analysis motives the reduction factor reported in section \ref{sec:results}.
\end{document}